\def\year{2022}\relax
\documentclass[letterpaper]{article} 
\usepackage{aaai22}  
\usepackage{times}  
\usepackage{helvet}  
\usepackage{courier}  
\usepackage[hyphens]{url}  
\usepackage{graphicx} 
\usepackage{natbib}  
\usepackage{caption} 
\DeclareCaptionStyle{ruled}{labelfont=normalfont,labelsep=colon,strut=off} 
\usepackage{algorithm}
\usepackage{algorithmic}

\usepackage{newfloat}
\usepackage{listings}
\floatstyle{ruled}
\newfloat{listing}{tb}{lst}{}
\floatname{listing}{Listing}

\usepackage{graphicx}
\usepackage{tabularx}
\usepackage{soul}
\usepackage{amssymb}
\usepackage{lscape}

\usepackage{physics}
\usepackage{amsmath}
\usepackage{tikz}
\usepackage{mathdots}
\usepackage{yhmath}

\usepackage{multirow}
\usepackage{comment}

\usepackage{subfigure}

\title{Memotion Analysis through the Lens of Joint Embedding (Student Abstract)}
\author {
    Nethra Gunti\textsuperscript{\rm 1*}, \quad
    Sathyanarayanan Ramamoorthy\textsuperscript{\rm 1*}, \quad
    Parth Patwa \textsuperscript{\rm 2}, \quad
    Amitava Das \textsuperscript{\rm 3, 4}
}
\affiliations {
     IIIT Sri City, India\textsuperscript{\rm 1} \quad \\
    UCLA, USA\textsuperscript{\rm 2} \quad \\
    Wipro AI Labs, India\textsuperscript{\rm 3} \quad \\
    AI Institute, University of South Carolina, USA\textsuperscript{\rm 4}
    \\
    \{nethra.g18, sathyanarayanan.r18\}@iiits.in, parthpatwa@ucla.edu, amitava.das2@wipro.com
}

\begin{document}

\maketitle

\begin{abstract}
   
Joint embedding (JE) is a way to encode multi-modal data into a vector space where text remains as the grounding key and other modalities like image are to be anchored with such keys. Meme is typically an image with embedded text onto it. Although, memes are commonly used for fun, they could also be used to spread hate and fake information. That along with its growing ubiquity over several social platforms has caused automatic analysis of memes to become a widespread topic of research. 
In this paper, we report our initial experiments on Memotion Analysis problem through joint embeddings. Results are marginally yielding SOTA.

\end{abstract}

\renewcommand{\thefootnote}{\fnsymbol{footnote}}
\footnotetext[1]{Equal contribution.}
\renewcommand*{\thefootnote}{\arabic{footnote}}
\setcounter{footnote}{0}

\section{Text with Image Joint Embeddings}

\begin{figure}[ht!]
\hfill
\subfigure[The king, queen analogy]{\includegraphics[width=3cm]{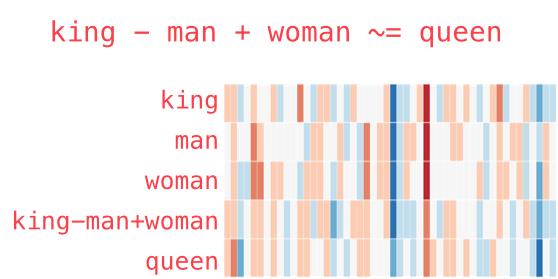}}
\label{fig:king_queen_word}
\hfill
\subfigure[An expected joint embedding(JE) space: king\textsubscript{JE}- queen\textsubscript{JE} $\approx$ \textit{boy}\textsubscript{JE} - girl\textsubscript{JE}]
{\includegraphics[width=4cm]{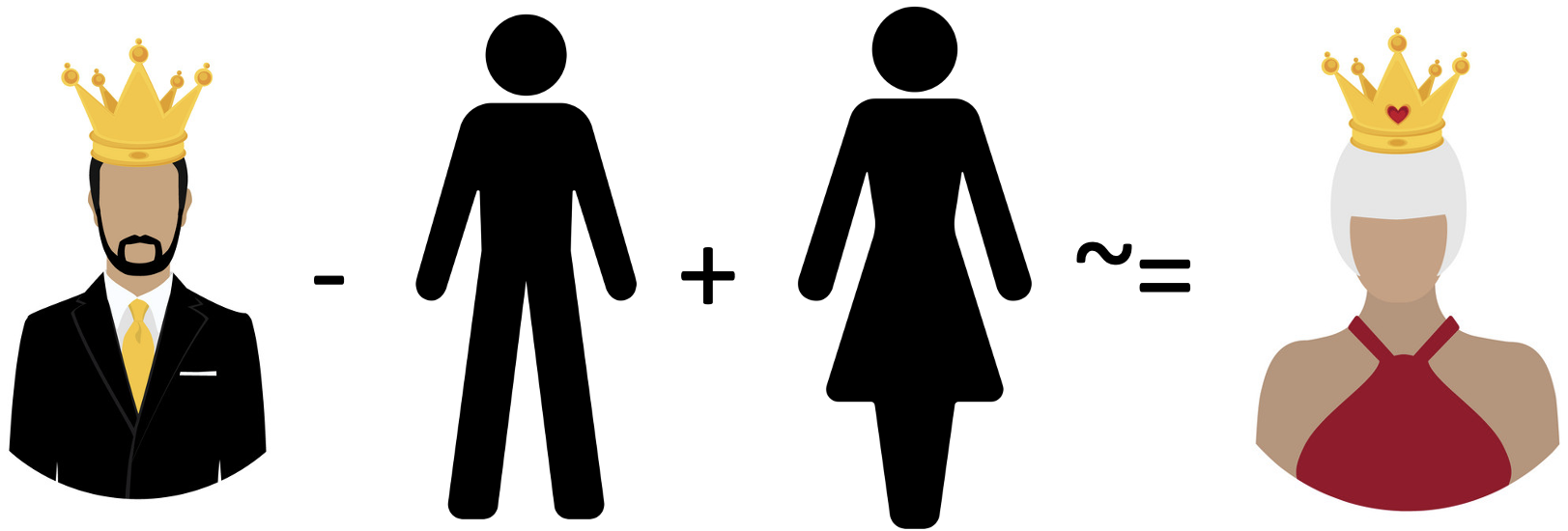}}
\hfill
\label{fig:king_queen_image}
\caption{ CNN based image embeddings are unable to capture contextuality like existing word embeddings.}
\label{fig:analogy}
\end{figure}

Distribution based compositional word embeddings like Word2vec, GloVe are popular in modern NLP. Consider the \textit{king, queen} word vector analogy (Fig.\ref{fig:analogy}.), which shows how good these word embeddings are at capturing syntactic and semantic regularities in language. However, CNN based image embeddings fail to capture such contextuality (Fig. \ref{fig:analogy}), since they are only source image dependant and do not capture corpus level distributions. Our work is focused towards textual grounding of images and to achieve distributional representation quite similar to word vectors. 

\section{Memotion Analysis - Data and SOTA}
We use the Memotion data \cite{sharma2020semeval}, consisting of 10K annotated memes, and OCR extracted text. Memotion analysis has 3 tasks- Task A: Sentiment Analysis (\textit{positive, negative, neutral}). Task B: Emotion Analysis (sarcasm, humour, offense, motivation). Task C: Semantic Class Prediction of emotional sub-classes. Reported SOTA (F1) results on these tasks are 35.47\%, 51.84\%, and 32.35\% respectively. (\cite{sharma2020semeval} reports details of SOTA). 

\section{Learning Joint Embedding}

Learning text+image JE has gained significant momentum recently. Majority of them are based on some form of Canonical Correlation Analysis (CCA), which finds similarities between two modalities by projecting them into one common vector space. We use two such existing models:

i) CLIP \cite{radford2021learning}, uses a visual transformer as the image encoder and a transformer as text encoder, to generate JE embeddings. It was pre-trained on a large dataset consisting of over 400M images.

ii) Stanford's Joint Embedding \cite{Kolluru2019ANA} (StanJE) uses VGG-19 and GLoVe to generate the image and text encodings respectively. They introduce a two-branch embedding network and a triplet loss function, which applies a margin based penalty, to obtain joint representation. It is pre-trained on Flickr30k \cite{plummer2016flickr30k} dataset.

These models are trained with sentence-image pairs (SJM). So we re-train both the models with word-image pairs. Further, we fine tune the models using Memotion data+Flickr30k data to obtain JE. 

\subsection{Word with Image Joint Embedding - (WJM)}

The smallest meaningful unit language is word. So, anchored word representations along with image encoding should be the expected milestone for JE. We re-train both the models with word-image pairs from Flickr30K and Memotion data. To get meaningful words, we apply choose only those words belonging to specific Parts-of-Speech (POS) \textit{noun, verb, adverb, adjective} categories. This results in a near-exponential increase in the dataset size, as every image is paired with multiple unique and significant words. Fig. \ref{fig:visualisation}, shows that the obtained JE space is able to capture meaningful analogy: boy\textsubscript{JE}- girl\textsubscript{JE} $\approx$ man\textsubscript{JE} - woman\textsubscript{JE}.

\begin{figure}
    \centering
    \includegraphics[width=\linewidth]{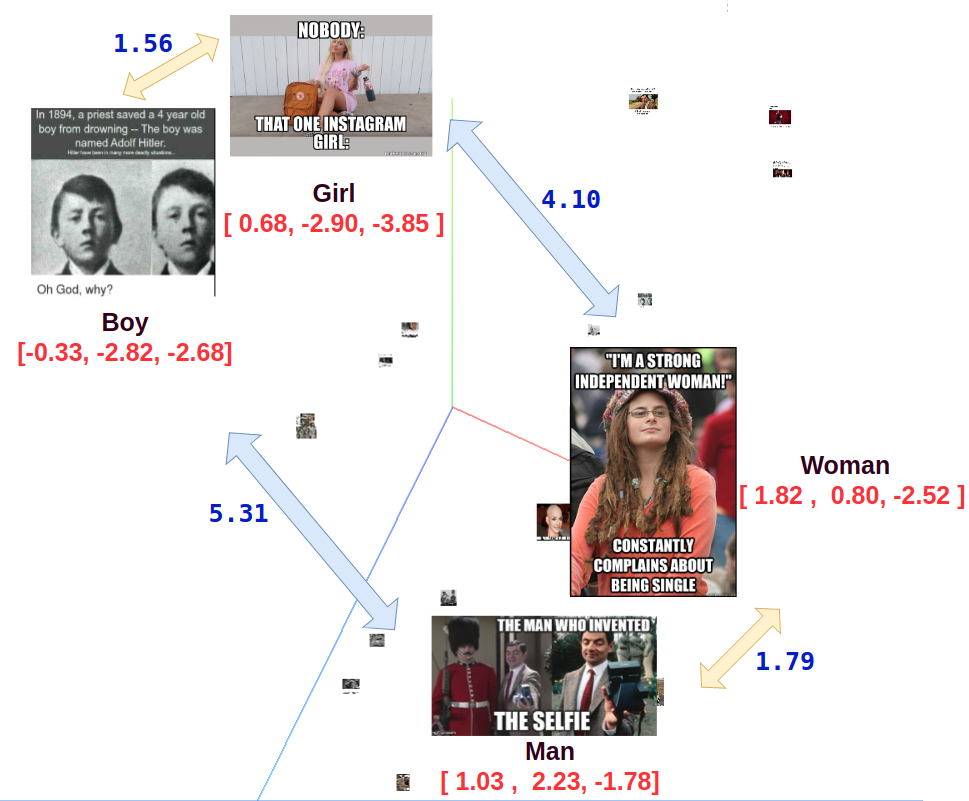}
    \caption{Visualising Analogy- emulating the king, queen analogy but instead using boy, girl, man, woman.  boy\textsubscript{JE}- girl\textsubscript{JE} $\approx$ man\textsubscript{JE} - woman\textsubscript{JE}.}
    \label{fig:my_label}
\end{figure}



\section{Memotion Analysis using Joint Embeddings}
Learning joint representations is a new research topic, similarly how to use such learned JEs for several downstream tasks - is an open territory for research exploration. CLIP allows zero-shot prediction and StanJE allows fine-tuning mechanism to use trained JEs for a particular task. We find that both the methods are not adequate for the Memotion task. Therefore, we propose new methods - Task A is a multi-class problem, so we use a Multi-Layer Perceptron (MLP) with 3 dense layers and a softmax output layer trained using categorical cross-entropy loss. For both the Task B and Task C, a Multi-Task Learning (MTL) framework, similar to \cite{safi-samghabadi-etal-2020-aggression}, is applied, which consists of four MLPs (one per sub-task). Binary cross-entropy loss is used for Task B and \textit{motivation} emotion of Task C, since they are binary prediction problems. Rest of the sub-tasks in Task C are multi-class problems, so we use categorical cross-entropy loss.
The code is available at \url{https://github.com/NethraGunti/Memotion-Analysis-Through-The-Lens-Of-Joint-Embedding}

\section{Results}
Table \ref{table:results} reports the performance of all the experiments. CLIP is a data hungry model, so it fails to perform well on datasets with less and diverse data. Hence, it performs poorly on Memotion Analysis tasks as compared to StanJE. WJM performs better than SJM in all the task. This validates our assumption that words are ideal choice for pairing with images for multimodal concept learning.  
The Word-Image Joint Embeddings obtained using the StanJE architecture yield the SOTA performance in all the categories.

\begin{table}[ht!]
\fontsize{9pt}{9pt}\selectfont
{%
\begin{tabular}{|c|c|c|c|}
\hline
Models & Data & CLIP & StanJE \\ \hline
\multicolumn{4}{|l|}{Sentiment Analysis- Macro-F1 Scores} \\ \hline
\multirow{3}{*}{SJM} & Flickr30k & 0.31 &  0.305 \\ \cline{2-4} 
 & Memotion & 0.294 & 0.311 \\ \cline{2-4} 
 & Combined & 0.3 & 0.284 \\ \hline
\multirow{3}{*}{WJM} & Flickr30k & 0.301 & 0.362 (+2\%) \\ \cline{2-4} 
 & Memotion & 0.31 & \textbf{0.373} (+5\%) \\ \cline{2-4} 
 & Combined & 0.307 & 0.354 \\ \hline
SOTA & \multicolumn{3}{c|}{0.355 \cite{sharma2020semeval}} \\ 
\hline
\multicolumn{4}{|c|}{Emotion Analysis- Macro-F1 Scores} \\ \hline
\multirow{3}{*}{SJM} & Flickr30k & 0.484 & 0.474 \\ \cline{2-4} 
 & Memotion & 0.47 & 0.484 \\ \cline{2-4} 
 & Combined & 0.476 & 0.423 \\ \cline{2-4} 
 \hline
\multirow{3}{*}{WJM} & Flickr30k & 0.481 & 0.484 \\ \cline{2-4} 
 & Memotion & 0.461 & \textbf{0.522} (+0.77\%) \\ \cline{2-4}
 & Combined & 0.453 & 0.455 \\ \hline
\multicolumn{1}{|l|}{SOTA} & \multicolumn{3}{l|}{0.518 \cite{sharma2020semeval}} \\
\hline

\multicolumn{4}{|c|}{Semantic Classification- Macro-F1 Scores} \\ \hline
\multirow{3}{*}{SJM} & Flickr30k & 0.282 & 0.238 \\ \cline{2-4} 
 & Memotion & 0.287 & 0.309 \\ \cline{2-4}
 & Combined &  0.236 & 0.309 \\
 \hline
\multirow{3}{*}{WJM} & Flickr30k & 0.281 & 0.261 \\ \cline{2-4} 
 & Memotion & 0.273 & \textbf{0.324}(+0.62\%) \\ \cline{2-4} 
 & Combined & 0.275 & 0.243 \\ \hline
\multicolumn{1}{|l|}{SOTA} & \multicolumn{3}{l|}{0.322 \cite{sharma2020semeval}} \\ \hline
\end{tabular}
}
\caption{Results of Memotion Analysis using JE.  
Reported results are of 5-fold cross validation.  
}

\label{table:results}
\end{table}

\section{Conclusion and Future Work}

In this work we report our initial experiments to solve Memotion analysis through JE. Results are marginally yielding SOTA. Our major contribution could be summarised as: (i) To the best of our knowledge this is the first endeavor to solve Memotion analysis through JE. ii) JE techniques have so far only been used for either image captioning task or text-based image retrieval task. Memotion analysis is arguably a much more complex multi-modal task.
(iii) We propose a different setup on how to use obtained JE to solve a multi-modal task. We carefully choose relatively simpler architectures like MLP and a simple version of MTL. The reason for such choices is to see and assess how useful JEs are - otherwise it could be argued that it is not the JE, but rather the architecture which is performing well. (iv) Results show word-image pairs are better alternatives than sentence-image pairs for JE training.
There are many challenges in working with JEs, For Eg: Image and Text associated with a meme are not necessarily related yet they convey reasonable meaning. Future work should be towards such considerations.

\fontsize{9pt}{10pt}\selectfont
\bibliography{aaai22.bib}

\end{document}